\documentclass[12pt]{article}
\usepackage{amsmath}
\usepackage{graphicx,psfrag,epsf}
\usepackage{enumerate}
\usepackage{url} 
\usepackage{array}
\usepackage{siunitx}
\usepackage{listings}
\usepackage{booktabs}

\usepackage{array}

\newcommand{\blind}{0}

\usepackage{setspace, amsmath, amssymb,  url, multirow, booktabs, verbatim, bm,threeparttable, color,subfig}
\usepackage{amsthm, rotating}
\usepackage{longtable}
\usepackage{float}
\usepackage{setspace}
\usepackage{helvet}
\usepackage{wrapfig,lipsum,booktabs}
\usepackage{ ulem }
\usepackage{xcolor}
\usepackage{algorithm}
\usepackage{algpseudocode}
\usepackage[round]{natbib}
\usepackage[letterpaper,margin=1in]{geometry}

\usepackage{graphicx} 
\usepackage{grffile}
\begin{document}

\def\spacingset#1{\renewcommand{\baselinestretch}%
{#1}\small\normalsize} \spacingset{1}


\if0\blind
{

\title{\bf Enhancing Functional Data Analysis with Sequential Neural Networks: Advantages and Comparative Study}
\author{Jianxiang Zhao\\
     Intelligent Information Processing Laboratory, \\Hangzhou Dianzi University, Hangzhou\\
    Jingjing Li \\
    Department of Mathematics and Statistics, \\Wake Forest University, Winston-Salem\\
    Minghan Chen \\
    Department of Computer Science, \\Wake Forest University, Winston-Salem\\
    and\\
    Sneha Jadhav\thanks{
    Corresponding author} \\
    Department of Mathematics and Statistics, \\Wake Forest University, Winston-Salem
}

  \maketitle
} \fi

\if1\blind
{
  \bigskip
  \bigskip
  \bigskip
  \begin{center}
    {\LARGE\bf Title}
\end{center}
  \medskip
} \fi

\bigskip
\begin{abstract}
Functional Data Analysis (FDA) is a statistical domain developed to handle functional data characterized by high dimensionality and complex data structures. Sequential Neural Networks (SNNs) are specialized neural networks capable of processing sequence data, a fundamental aspect of functional data. Despite their great flexibility in modeling functional data, SNNs have been inadequately employed in the FDA community. One notable advantage of SNNs is the ease of implementation, making them accessible to a broad audience beyond academia. Conversely, FDA-based methodologies present challenges, particularly for practitioners outside the field, due to their intricate complexity. In light of this, we propose utilizing SNNs in FDA applications and demonstrate their effectiveness through comparative analyses against popular FDA regression models based on numerical experiments and real-world data analysis. SNN architectures allow us to surpass the limitations of traditional FDA methods, offering scalability, flexibility, and improved analytical performance. Our findings highlight the potential of SNN-based methodologies as powerful tools for data applications involving functional data.
\end{abstract}

\noindent%
\vfill

\newpage
\spacingset{1.45} 
\section{Introduction}
\label{sec:intro}

Advancing technology has enabled the collection of substantial data that are ordered in time, space, and other domains, such as wearable devices tracking physical activity, blood oxygen level, heart rate at regular time intervals, and house prices organized by region (space). Multivariate methods although useful do not take into consideration the main feature of data, i.e., its ordered/sequential nature. Moreover, correlations in the data pose additional challenges to multivariate methods. Two approaches, Functional Data Analysis (FDA) and Sequential Neural Networks (SNNs), have been developed for such data, each with its own advantages and drawbacks in both theoretical and practical aspects. The two approaches are currently studied independently as far as we know. Researchers tend to use one or the other method based on their backgrounds. For example, in the FDA community, SNNs are rarely considered as alternative methods for benchmarking and comparison. Given that such ordered and correlated data arise in crucial applications such as health, we bring attention to both approaches as well as compare and contrast them. In the process, we also provide comprehensive and high-level knowledge of both approaches, which will allow a wider audience to access and leverage their potential benefits.

\begin{figure}
	\begin{center}
		\includegraphics [width=0.8\textwidth]{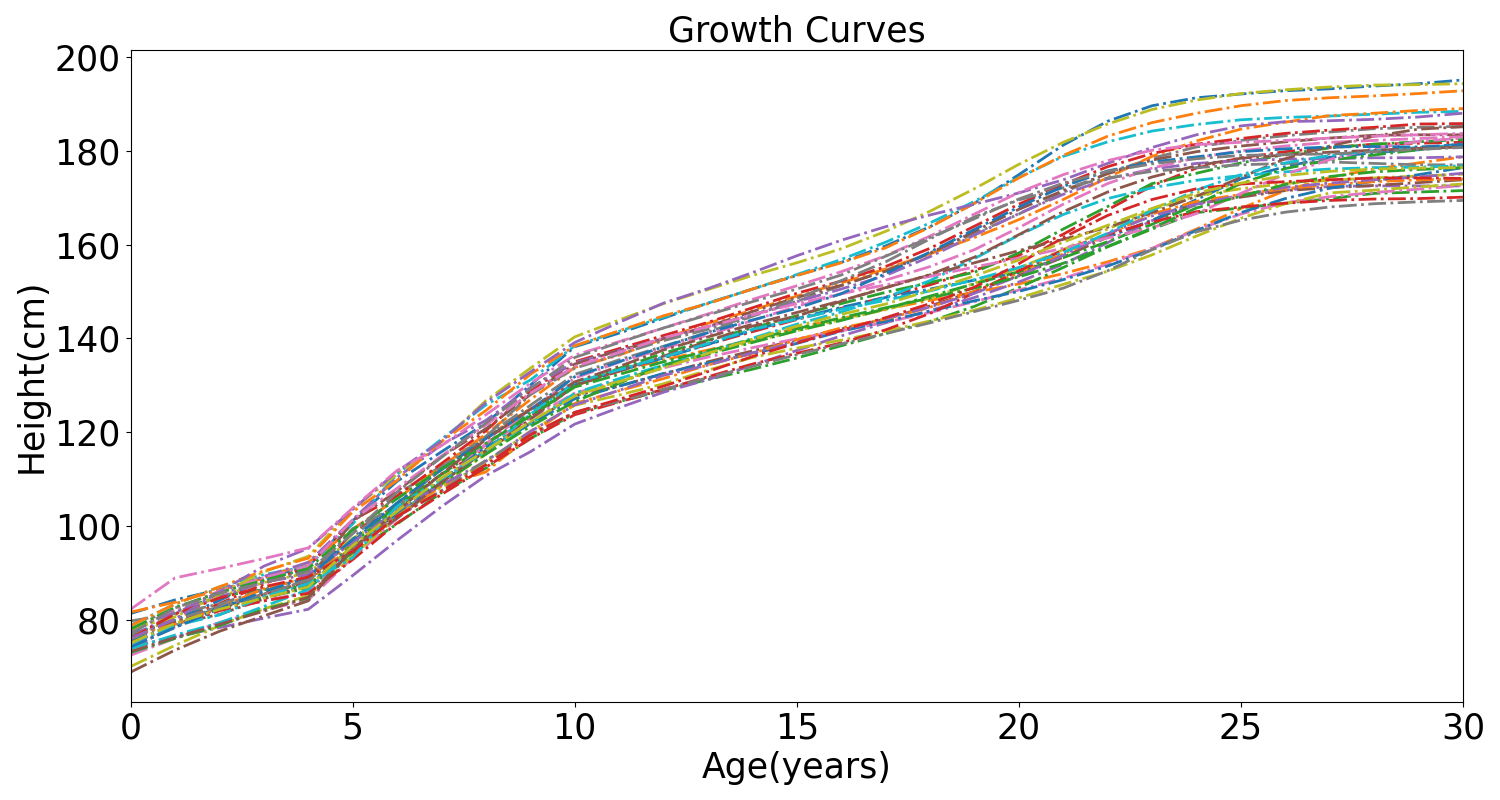}
	\end{center}
	\caption{Berkeley Age Study: children's heights at different ages. \label{berk}}
\end{figure}

Often, plots of sequential data reveal underlying patterns that can be interpreted as realizations of functions or curves. FDA has evolved over the past two decades to take advantage of functional characteristics such as smoothness, shapes, and derivatives among others \citep{wang2016functional}. FDA-based methods treat the collection of observations at different time points as a single function. This provides a more natural, informative, and holistic approach to such data. Consider the Berkely growth data containing heights of children at different ages in Figure \ref{berk}. Treating the entire data from a child as a realization of a continuous function allows us to study the rate of growth, acceleration, the onset of puberty, etc. There has been an enormous growth of FDA-based methods, offering a plethora of insights that might not be attainable through multivariate techniques. Of these, functional regression models have gained considerable popularity. Regression models can be used to make predictions or determine relationships between variables in diverse fields, such as health and finance. 
Though theoretical complexities make functional regression obscure to practitioners outside of the statistical or mathematical domains, many popular FDA methods can be applied with the help of several easy-to-use software packages or codes provided by authors \citep{morris2015functional}. However, even with these aids, unless a practitioner possesses a profound understanding of FDA and can precisely identify the most suitable model from the array of available options, they often find themselves experimenting with various approaches. This need for extensive knowledge in FDA poses a formidable barrier when working with ordered data sets.

Sequential Neural Networks that use recurrent connections such as Recurrent Neural Network (RNN) \citep{rumelhart1985learning}, Long Short Term Memory (LSTM) \citep{hochreiter1997long}, and Gated Recurrent Unit (GRU) \citep{goodfellow2016deep}, are classical go-to approaches in machine learning for analyzing time-series, image, text, and other types of datasets that have inherent sequential structures. More recently, attention models \citep{bahdanau2014neural,niu2021review,vaswani2017attention} have become popular given their improved performance in handling sequential data. These models possess the capability to dynamically regulate the attention or weight assigned to inputs at different times according to their importance and relevance. Massive attention models with millions of parameters have revolutionized natural language processing. Functional data that is clearly sequential in nature typically has a much simpler structure than text data. We explore the viability of using relatively simpler SNN models such as RNN or LSTM. Though neural network based models are not known to provide interpretable results, they are known for their prediction power, particularly in the presence of large data sets. Despite having complexities on both the theoretical and computational front, it may be relatively easier to understand the basics of SNN architecture for a meaningful implementation compared to the FDA-based regression methods. The network architecture of neural networks may vary according to the task, but the core components (weights, biases, forward propagation, optimization) remain largely unchanged making them readily adaptable for various tasks; while in the domain of statistics and machine learning, new methods along with their appropriate models, estimations, and implementation details must be developed from scratch. On the computational side, the immense advances in deep learning packages facilitate the easy implementation of different neural network architectures. The same cannot be said for FDA methodologies as they often require different packages, software, input/output formats, and most importantly a reasonable understanding of the methodology. The implementation and interpretation of SNNs are straightforward, which motivates the exploration of SNNs for functional data analysis and comparison with FDA-based approaches. We provide brief explanations and develop simulation studies for a comparison of the two approaches. We demonstrate that SNNs not only exhibit comparable or superior performance, but also provide a highly versatile platform to fit different types of linear, non-linear models with varying dimensions of the input/output unlike FDA-based methods.

{The rest of this paper is organized as follows. Sections 2 and 3 introduce FDA regression methods and sequential neural networks, respectively. We compare their performance in various simulation settings in Section 4, and then apply them to real functional data in Section 5. In Section 5, we conclude with some remarks and future directions. Our focus is on the basic and common practices of both approaches, to examine if SNNs can be employed for functional data.}



\section{Functional Data Analysis}

{In Functional Data Analysis (FDA), data transformation and representation play a crucial role as they enable the conversion of raw discrete data points into smooth function curves or finite-dimensional coefficient vectors. This facilitates subsequent data analysis and modeling processes. Data analysis and modeling constitute the core steps of FDA, wherein appropriate functional regression models can be selected to describe the relationship between the output and input variables. Additionally, methods such as Principal Component Analysis, Functional Analysis of Variance, and others can be utilized for data dimensionality reduction and feature extraction, thereby revealing the underlying structures and patterns within functional data. }

\subsection{Transformation from Discrete Data Points to Smooth Functions}
{FDA offers methods for analyzing data containing functions. However, obtaining a complete function curve is not feasible; only discrete data points can be observed. Many approaches utilize these discrete observations to recover the underlying function, based on the assumption that the observed values exhibit typical smoothness in their discrete implementation. Nevertheless, the observed data is often subject to measurement errors, leading to deviations from the underlying function curve. Consequently, smoothing plays a crucial role in FDA by transforming the raw discrete data points into a smoothly varying function curve, with the aim of minimizing short-term deviations caused by observation errors such as measurement errors or system noise, while highlighting patterns within the data.}

Suppose, we observe data $Y_1,...,Y_T$, at time points $t_1,...,t_T$. Let $Y(\cdot)$ denote the underlying function, i.e., $Y(t_j)=Y_j, \forall j.$  Basis expansion postulates that $Y(t) = \sum_{j=1}^{\infty} \theta_j\phi_j(t),$ where $\phi_j(t), j \geq 1$ are basis functions. In practice, functions can be well approximated using a small or finite number of basis functions, i.e., $Y(t) = \sum_{j=1}^{q} \theta_j\phi_j(t).$ Thus, for a selected value of $q$ and basis functions, we obtain the model $Y_j = \sum_{i=1}^{q} \theta_i \phi_j(t_j), j=1,...,T.$ This is a regression model with observed data as response and basis function as predictors, allowing us to readily obtain estimates $\widehat{\theta}_1,...,\widehat{\theta}_q$. The function underlying the observation is approximated as $ \sum_{j=1}^{q} \widehat{\theta}_j\phi_j(t).$ This is referred to as a smoothing technique. More advanced versions of the technique using penalization can be found in \cite{ramsay2008functional}. Several examples of FDA can be found in \cite{ramsay2008functional}.  The function passing through the data in Figure \ref{berk} can be obtained via such a smoothing technique. Many methodologies in FDA preprocess data to obtain underlying functions before conducting further analysis.
%

\subsection{Selection and Challenges of Functional Regression Models}

{
Functional regression models, similar to their multivariate counterparts, are determined by factors such as:
i) The nature of the dependent (output) variable, which can be discrete or continuous.
ii) The input (independent variable) output relationship, which can be linear, non-linear, parametric, non-parametric, or semi-parametric.
Regularization and correlated samples add another dimension that we do not consider here. Most of these methods can be easily implemented using packages in software such as R, SAS, etc. For functional regression models, in addition to these factors, we also need to consider the following factors.}


{(1) Functional Variables--A functional linear model with both input and output functions differs from a model with only an input or a function.
(2) Density/Quantity of Discrete Observations--Dense data can be easily recovered from underlying functions, while sparse data presents challenges.
(3) Temporal Dependency between Input and Output. For example, to study the relationship between step count and activity intensity, we can examine the relationship between step count at time $t$ and intensity at the same time. To model the relationship between daily activity and BMI, we can consider the entire activity function as the input. Additionally, historical models can be considered, where the output at time $t$ depends on the entire input history up to that time.}

{Even after determining all these components, the treatment of functional data can vary. Functions exist in an infinite-dimensional space, which means that functional regression involves estimating parameters that are also functions. Consequently, functional regression becomes an ill-posed regression problem. To address this issue, a commonly employed approach is a combination of spectral decomposition and truncation. Previous works have utilized regularization techniques to tackle this challenge \citep{hall2007methodology,chakraborty2016hybrid}. We have only highlighted a few factors that need to be considered when selecting a functional regression model. In some cases, multiple models exist for the same problem, each with its own set of assumptions and estimation methods. While this plethora of choices increases the likelihood of finding a model that meets the data requirements, it also necessitates familiarity with functional data literature. Although certain methods are accompanied by user-friendly software, this is not always the case, making it challenging to explore various options for analyzing a dataset. This is where neural networks possess an advantage, as they provide a unified platform for implementing multiple neural network architectures. In the following section, we will provide a brief overview of some widely used function regression architectures. }

\subsection{Functional Predictor Regression Models}
These models have a scalar response $Y$ and a functional regressor $X(t), t \in \mathcal{I}$. The linear model is written as 
    \begin{align}  \label{flm}
    Y = \int_{\mathcal{I}} X(t)\beta(t) +e,       
    \end{align}
where $\beta(\cdot)$ is the model parameter function and $e$ is the model error. Clearly, this model can be viewed as a generalization of the multivariate regression model, where the sum is replaced by the integral. Asymptotics and estimation procedures for this model have been developed extensively under varying frameworks \citep{wang2016functional,cardot2003spline,cai2012minimax}. Model \eqref{flm} is a functional version of the linear regression model. However, when dealing with binary data, more sophisticated models are necessary, such as the semi-parametric model defined as:
\begin{align}  \label{non3}
Y = g\left(\int_{\mathcal{I}} X(t)\beta(t)\right)+e. 
\end{align}  
There have been developments in the semi-parametric with known link functions \citep{james2002generalized,cardot2005estimation,dou2012estimation} as well as unknown link functions \citep{chen2011single,muller2005generalized}. Furthermore, there is the non-parametric model, formulated as 
\begin{align}  \label{non1}
 E(Y | X) = r(X),   
 \end{align}  
where $r$ is an operator estimated using functional kernel \citep{ferraty2006nonparametric}. Other types of models are also prevalent, such as additive, single index ones \citep{mclean2014functional, muller2008functional,hall2007methodology,chen2011single}.  Regularization is another crucial factor influencing model considerations \citep{chakraborty2016hybrid}. In the initial stages of data analysis, it is often challenging to determine which model is best suited for the data. As a result, exploring several different models becomes essential, whenever possible. In Table \ref{tab:R}, we provide a list of packages available in R that can be utilized to implement scalar functional regression models. Additionally, we outline the general categories of models associated with these packages. We have also listed the broad category of their model. Note that even if the model type is the same (e.g., linear), the actual model, estimation methods, or other details may differ between different methods represented by different functions. Further resources regarding software can be found in \cite{morris2015functional}.

\begin{table}[h]
\centering
\caption{List of R packages used for function regression models.}
\label{tab:R}
\begin{tabular}{ lll } 
 \toprule 
 \multicolumn{1}{c}{\textbf{R Package}} & \multicolumn{1}{c}{\textbf{Function name}} & \multicolumn{1}{c}{\textbf{Model Type}}\\
 \midrule 
 \texttt{fda} & \texttt{fRegress} & linear model \\ 
 \addlinespace 
 \texttt{fda.usc} & \texttt{fregre.basis, fregre.bootstrap, fregre.lm} & linear model\\ 
 \addlinespace
 \texttt{fda.usc} & \texttt{fregre.gkam, fregre.glm, fregre.gsam} & linear model\\ 
 \addlinespace
 \texttt{fda.usc} & \texttt{fregre.np} & non-parametric model\\ 
 \addlinespace
 \texttt{fdapace} & \texttt{FLM} & linear model \\ 
 \addlinespace
 \texttt{fdapace} & \texttt{FAM} & additive model\\ 
 \addlinespace
 \texttt{refund} & \texttt{fgam} & additive model\\ 
 \addlinespace
 \texttt{refund} & \texttt{fpcr} & generalized linear model\\ 
 \addlinespace
 \texttt{fdamixed} & \texttt{fdaLm} & linear mixed-effects model\\
 \bottomrule 
\end{tabular}
\end{table}


\subsection{Function on Function Regression Models}

In this scenario, both the response denoted as $Y(t), t \in \mathcal{I}$ and the regressor $X(t), t \in \mathcal{I}$ are functions. There are several types of models here dictated by the time aspect relationship of the input to the output. The concurrent model uses the input at time $t$ to predict the output at the same time $t$. Let $e(\cdot)$ denote error function and  $\alpha(\cdot),~\beta(\cdot)$ denote parameters functions.  The concurrent linear model is given as
\begin{align}
Y(t) = \alpha(t) + \beta(t) X(t) +e(t).
\end{align}
This is a basic model structure with multiple methods available to estimate the model parameters \citep{ramsay2008functional, manrique2018ridge, wu1998asymptotic,csenturk2010functional}.  There are models that allow more general non-linear relationships with varying degrees of complexity and flexibility. We give three models as examples \citep{maity2016general,scheipl2015functional,scheipl2016generalized}, shown below.
\begin{align}
&Y(t) = \alpha(t) +f(X(t))+e(t),\\
&Y(t) = \alpha(t) +F\{X(t),t\}+e(t),\\
&g(E(Y(t)|X(t),t))=  F\{X(t),t\},
\end{align}
where $g(\cdot)$ is a known link function while the remaining functions are unknown.  A large number of regressor functions might call for variable selection  \citep{goldsmith2017variable}. The non-concurrent model uses the entire input or in some cases a part of it to predict $Y(t)$. The most common non-concurrent model is the following linear model.
\begin{align}\label{m2}
Y(t) = \alpha(t) + \int_{s \in \mathcal{I}} \beta(t,s) X(s)ds +e(t).
\end{align}

Parameter estimation of this model has been addressed in many works, some of which are \cite{ramsay1991some,wu2011response,besse1996approximation,cuevas2002linear,bernardi2021locally}. The method in \cite{yao2005functional} suggests estimation for the case when data is sparsely observed. A model based on Tikhonov regularization is proposed in \cite{benatia2017functional}. While the literature primarily focuses on linear models, there are also works that consider non-linear cases.
One such model from \cite{scheipl2016generalized} is 
\begin{align}
Y(t) = \int f(X(s),s,t)ds + e(t).
\end{align}
Some other non-linear models can be found in \cite{muller2008functional,ma2016dynamic}. A wide range of models is accessible when both input and output are functions \citep{morris2015functional}. There are also historical models where past values of the data are used to predict the current value \citep{malfait2003historical, morris2015functional}. 
For implementing the non-concurrent linear model, several options are available.
One can use \textit{fregre.basis.fr} function from R package \textit{fda.usc}, \textit{FLM} function from library \textit{FDApace}, or \textit{pffr} from library \textit{refund}. The latter two offer easier implementations as they mainly require data input from the user. The \textit{pffr} function can be used easily to fit concurrent, non-concurrent, linear, and non-linear models. It also offers options to incorporate random effects. 

\subsection{Functional Response Regression Models}

This framework is relatively less explored compared to the previous ones. In this case, the only functional variable is the response. A typical model is 
\begin{align}
Y(t) = \alpha(t) + X\beta(t) + e(t).
\end{align}

To estimate the effect $\beta(t)$ of $X$ on $Y(t)$, a simple and commonly used strategy suggested in \cite{ramsay2008functional} is to use the basis representation, i.e., $\beta(t) = \sum_{j=1}^{q}  b_j\phi_j(t)$. Least squares along with penalization are used to obtain regularized estimates. Mixed model approaches in \cite{morris2006wavelet,antoniadis2007estimation} treat the individual curves as random. More options for estimating the parameters can be found in \cite{reiss2010fast, chiou2004functional}.


\section{Sequential Neural Networks}

\begin{figure}[h]
	\begin{center}
		\includegraphics[width=6in, height=3in]{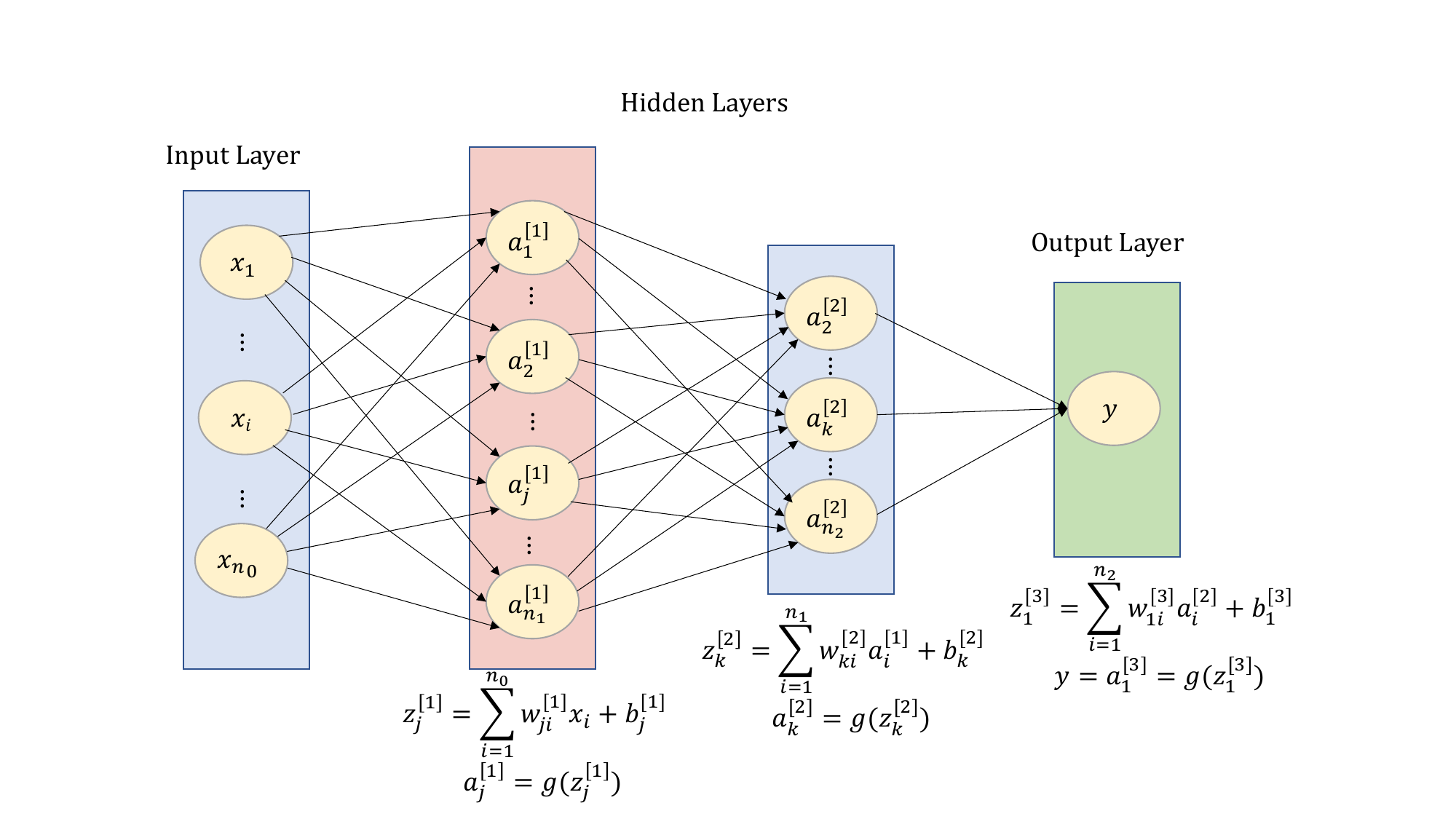}
	\end{center}
	\caption{A feed forward neural network with two hidden layers. \label{FNN}}
\end{figure}

Neural networks are much like lego blocks, different arrangements of the same blocks yield different architectures that can capture different data representations. This flexibility is very convenient for data applications.  Basic neural network blocks mainly consist of nodes, layers, and activation functions. Figure \ref{FNN} represents a basic neural network with an input layer, two hidden layers, and an output layer. The nodes in a layer (circles) represent mathematical operations and the arrows represent the inputs for these operations. Each node in a layer takes a weighted linear combination of the input or the output of the previous layer, adds a bias term, and passes the result ($Z_j^{[i]}$) to an activation function ($g_i$) that is typically non-linear. The number of outputs of each layer is determined by its number of nodes. The output of a neural network is obtained using compositions of several functions. A typical feed forward neural network with input $x\in R^d$ and $L$ layers  is represented as follows

\begin{align}
f(x,\theta) = g_L\Bigl[W_Lg_{L-1}\Bigl(W_{L-1}\ldots g_2\Bigl(W_2\bigl(g_1(W_1x+b_1)+b_2\bigr)\Bigr)+b_L\Bigr) \Bigr].
\end{align}

where the $\theta=(W_1,...,W_L,b_1,...b_L), W_L \in R^{d_l \times d_{l-1}}, d_0 =d, g_l: R^{d_l} \to R^{d_l}$ are known or predetermined activation functions \citep{bartlett2021deep}.  Each layer $i$, transforms the output of the previous layer as $ a_i = g_i(W_ia_{i-1}+b_i )$ and then feeds it to the next layer. Hence, the name ``feed forward neural network". The function $g_i(\cdot)$ is implemented component wise. The activation function determines the nature of the outputs of each layer. Thus, we can use a sigmoid activation function for a binary classification problem or a non-linear activation for the regression prediction problem. Thus, neural networks can be easily adapted to any type of output including multiple outputs with the help of activation functions and the number of nodes in the output layer. The weights ($W_i$) and biases ($b_i
$) are obtained to minimize a loss function that is determined by the problem at hand.  For example, mean squared error loss is used for a continuous output or cross-entropy for binary output. With enough data, neural networks have been found to work well in obtaining predictions for many complex data sets \citep{bartlett2021deep}. It is common for neural networks to involve a large number of parameters that may cause an overfitting issue. Though there are methods to avoid overfitting \citep{goodfellow2016deep} neural networks typically involve much larger models compared to statistical methods.  Even though neural networks use gradient descent based methods to solve highly non-convex problems they have a very high predictive accuracy. In \cite{bartlett2021deep}, this over parametrization is referred to as benign overfitting.

\begin{figure}
	\begin{center}
		\includegraphics[width=0.8\textwidth]{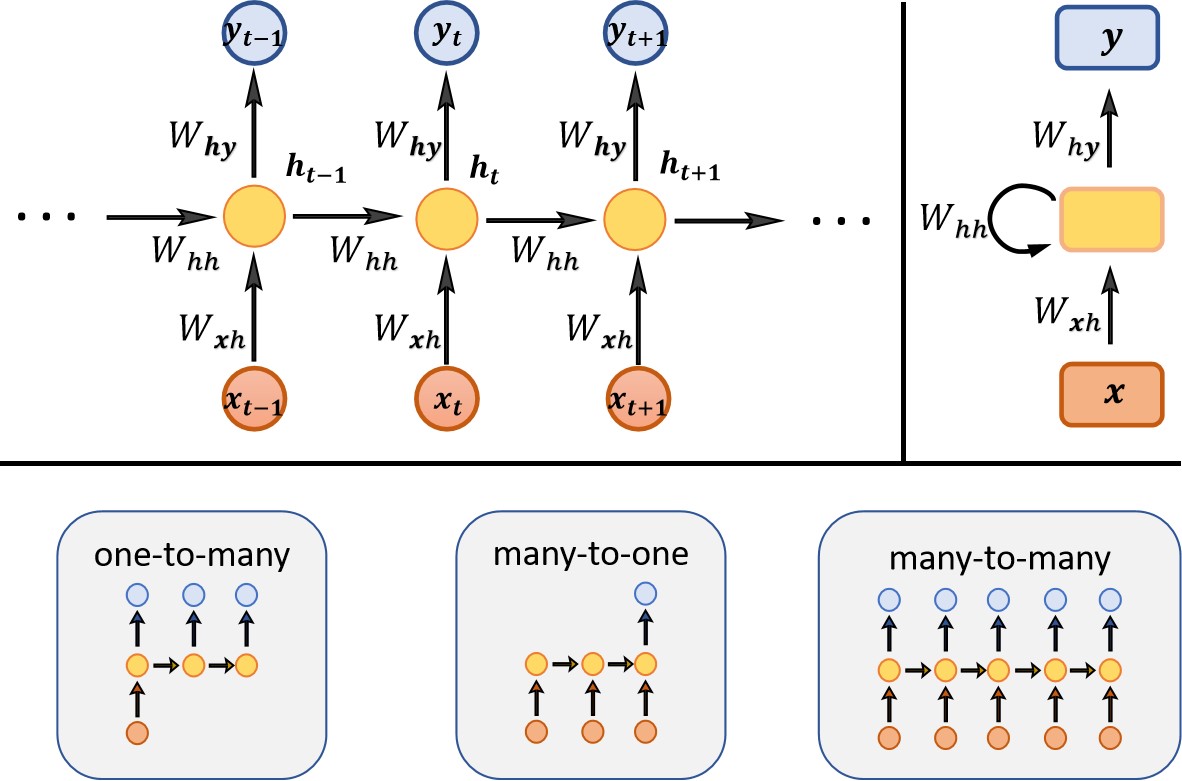}
	\end{center}
	\caption{Top left: A simple recurrent neural network (RNN) with common weight at the different time steps. Top right: A different representation of the same RNN using recurrent connection. Bottom: Different options for inputs/outputs in RNNs. \label{SNN}}
\end{figure}

These feed forward network models are not appropriate for data that has a sequential structure such as time series, functional data, or even text data as they do not have a mechanism to consider the order of the data. Sequential neural networks are a special type of neural networks designed for sequential data. Their main feature is that they propagate information through ‘time’ with units that have common weights (parameters). Equipped with self connections, multiple layers, and units, SNNs are able to capture long term dependencies in the data. {Figure \ref{SNN} shows a basic SNN model, namely an RNN that adapts to the sequential data structure. The notations along the arrows denote the weight matrices. For example, $W_{xh}$ is the weight matrix involved between input and the hidden layer. Note that the weights do not change over different time points, thus optimal weights capture the time changing relation between the inputs and outputs. Moreover, SNN can undergo different morphological changes according to the different task requirements. These include one-to-many, many-to-one and many-to-many.} There are other types of SNN such as LSTM \citep{hochreiter1997long}, GRU \citep{cho2014learning}, etc. While RNN suffers from short term memory \citep{bengio1994learning} and does not work when the length of the input sequence is large, LSTM and GRU overcome this issue to an extent. Given their advantages, neural networks have been used to model functional data \citep{conan2005representation,wang2020estimation,conan2002multi}. These works propose a feed forward type of functional architecture, where the input is a function, and the rest of the architecture resembles the ‘usual’ feed forward networks. Since it is not practical to input a function directly, these methods use a finite-dimensional representation of functions. This approach considers the entire functional input as a single entity, unlike an SNN that considers the functional data as a series of inputs observed sequentially. Moreover, these are suitable when the input is a function whereas the output is not a function. In this work, we examine the use of SNNs for prediction involving functional data input, output, or both. In the process, the flexibility of SNNs to model different types of input/output relationships becomes evident. 

Importantly, we are not constrained by restrictive models as in functional regression. It is easy to change network architecture to model different relationships. We can fit fairly complex models using high-level but easy-to-use libraries such as Keras, Tensorflow, etc. For example, Table \ref{tabcode} shows a code snippet of fitting the SNN model using Keras. First, we establish a sequential model. Then, we introduce an LSTM layer into the model to facilitate the processing of sequential data. Next, we add a dense layer to the model for predicting output results. After adding the layers, we continue to compile the model, specifying the optimizer, loss function, and evaluation metrics, among other parameters. Finally, we start the model training process, in which we input the training data (including input and output) into the model and perform backpropagation optimization based on the specified optimizer and loss function. This iterative process aims to gradually converge the model and obtain suitable parameter configurations.



\begin{table}[h]
\centering
\caption{Code snippet of SNN using Keras.}
\begin{tabular}{c ll}
\toprule
\label{tabcode}
\textbf{Step} & {\textbf{Code}} & {\textbf{Explain}} \\
\midrule
1 & \texttt{model = Sequential()} & Create a Sequential model \\
2 & \texttt{model.add(LSTM)} & Add a LSTM layer to the model  \\
3 & \texttt{model.add(Dense)} & Add a Dense layer to the model \\
4 & \texttt{model.compile(loss, optimizer, metrics)} & Compile the model \\
5 & \texttt{model.fit(input, label)} & Train the model \\
\bottomrule
\end{tabular}
\end{table}

It can be seen that one advantage of using SNNs for functional data analysis is that they are easy to implement with existing libraries and frameworks. These libraries provide high-level APIs that allow users to define, train and evaluate neural network models with minimal coding. Moreover, these libraries support multiple input/output scenarios, such as scalar-to-function, function-to-scalar, function-to-function etc. This means that users can easily adapt their models to different types of functional data analysis problems, without having to change the underlying architecture or algorithm. Deep learning frameworks are difficult to customize, but they have a clear computational advantage in dealing with large-scale data and complex models, compared to traditional statistical methods. For example, to customize an LSTM layer in the Keras framework, one not only needs to master the principles and mathematical formulas of LSTM, but also needs to use TensorFlow’s low-level API to implement them, which undoubtedly increases the complexity and time consumption, and may reduce the performance and compatibility. However, with the continuous development of artificial intelligence, deep learning frameworks also provide a rich and diverse set of APIs, which enable us to easily build the desired models without too much customization.

\section{Numerical Simulation}
\label{sec:verify}

{We conducted a comparative study of sequential neural networks and functional data analysis methods for different input functions and output scalars or functions. By generating functional data with linear and non-linear relationships, we employed various neural network models and functional regression models for prediction. These experiments were kept simple to control the number of factors that might influence the output, which helped avoid ambiguity and enhance clarity.}

\subsection{Scalar Response and Functional Predictor}

\subsubsection{Continuous Scalar Response}
    
For the experimental setup, the input functional data is generated as 
$X_{ij}=X_{i} (t_j )=\Sigma_{k=1}^q x_{ik} \phi_k (t_j ),   x_{ij} \sim N(0,0.5),i=1,...,n, t_j \in[0,1],j=1,...,100,$
and $\phi_k (\cdot)$ represents Fourier basis functions. Two cases are adopted to generate the response: 

\begin{itemize}
\item Linear Case: $Y_i= \int X_i (t)\beta(t)dt+e_i, $
\item  Non-linear Case: $Y_i = sin(\int  X_i (t)\beta(t)dt)+e_i,  $
 \end{itemize}
where $ e_i \sim N(0,0.5), \beta(t_j )=\Sigma_{k=1}^{q}  (1/k)  \phi_k (t_j )$. 

We generate a training set of size $n=200$ and a test set of size $n=50$. Neural networks are typically used for complex data sets and accordingly large sample sizes are required to train them. We deliberately choose a smaller sample size to examine its ``worst-case performance". Neural networks with recurrence such as RNNs and even LSTMs have a memory problem, i.e., they forget earlier data as the input sequence gets large causing their performance to deteriorate. With this in mind, we explore their performance when the input length is considerably large at 100. The number of basis functions $q$ determines the complexity of functions with larger values resulting in more complex functions. We vary data complexity to investigate its effect.
The following two neural network models are considered.

\textbf{Sequential Neural Network (SNN)}: This paper uses the original LSTM layer (Figure \ref{fig:SNN_model}) as the core component of the SNN, which can effectively solve the long-term dependency problem. Through the Many-to-Many structure, LSTM can not only extract patterns and rules from past historical information, but also predict future trends. This paper constructs a two-layer SNN model, the first layer is the original LSTM layer, containing 32 units, used to capture all the information of the sequence; the second layer is a hidden layer, used to produce the required format of output. The SNN model proposed in this paper is relatively small, but has high prediction accuracy.

\textbf{Feed forward Neural Network (FNN)}: This model treats all observations at different time points as separate variables, which are then fed into a simple feed forward neural network (Figure \ref{fig:FNN_model}). We use two dense hidden layers, each with 32 units, and then use another dense layer to generate the appropriate output.

We consider a variety of FDA approaches that are readily available in R. We examine functional linear models \eqref{flm}, denoted as FLM1  \citep{goldsmith2011penalized} and FLM2 \citep{yao2005functional}. FLM1 is implemented using \textit{pfr} function from \textit{refund} library, while FLM2 is implemented using \textit{FLM} function from R package \textit{fdapace}. FNLM denotes functional non-linear model  \eqref{non1} that uses kernel estimation from \cite{ferraty2006nonparametric}. This is implemented using \textit{fregre.np} function from \textit{fda.usc} library. FAM is the functional additive model from \cite{scheipl2016generalized} executed using \textit{pfr} function from \textit{refund} library.

The mean squared error for all the methods based on 500 replicates is reported in Table \ref{tab1}. In the case of simple data (where $q$ represents the complexity of the data), the functional regression model achieves the minimum error. FNN has larger errors than SNN, thus making a case for the importance of accounting for the sequential nature of the data. Overall, SNN demonstrated its superiority in handling complex data, achieving or surpassing the predictive performance of the FDA-based model. We observed that as the data complexity increased ($q$ increased), the performance of the FDA-based method deteriorated drastically, while the performance of SNN declined more mildly. Particularly in high-complexity scenarios, SNN outperformed the FDA-based model significantly. The drop is not as much for SNN. Moreover, the SNN model we use is quite small and relatively arbitrary so we do expect its performance to drop with increasing data complexity. However, its performance is easy to improve with architecture modifications. While the performance of FDA methods can be improved, it is less straightforward and simple compared to neural networks.

\begin{table}[h]
	\caption{Prediction errors (mean squared errors) for continuous scalar response.}
	\label{tab1}
	\begin{center}
		\begin{tabular}{ll | rrrrrr}
			\hline
			$q$	&	type	&	SNN	&	FNN	&	FLM1	&	FLM2	&	FNLM	&	FAM	\\
			\hline
			5	&	linear	&	1.39E-02	&	1.91E-02	&	4.70E-01	&	1.90E-02	&	9.47E-03	&	\textbf{9.45E-03}	\\
			9	&	linear	&	\textbf{1.61E-02}	&	1.97E-02	&	4.53E-01	&	2.00E-02	&	2.66E-02	&	2.62E-02	\\
			21	&	linear	&	\textbf{2.69E-02}	&	2.84E-02	&	4.53E-01	&	2.92E-02	&	6.74E-02	&	6.63E-02	\\
			5	&	sin	&	1.42E-02	&	1.89E-02	&	4.52E-01	&	1.91E-02	&	9.62E-03	&	\textbf{9.60E-03}	\\
			9	&	sin	&	\textbf{1.57E-02}	&	1.94E-02	&	4.37E-01	&	1.96E-02	&	2.33E-02	&	2.28E-02	\\
			21	&	sin	&	\textbf{2.18E-02}	&	2.43E-02	&	4.40E-01	&	2.47E-02	&	6.73E-02	&	6.55E-02	\\\hline
		\end{tabular}
	\end{center}
\end{table}

\subsubsection{Binary Scalar Response}

The functional inputs are generated exactly as in Section 3.1.1. We generate the binary output using the following two cases.

\begin{itemize}
\item Linear Case: $P(Y=1)= 1/(1+e^{-p}), p = \int X_i (t)\beta(t)dt $
\item Non-linear Case: $P(Y=1)= 1/(1+e^{-p}), p = sin(\int X_i (t)\beta(t)dt )$
\end{itemize}

The simulation settings are the same as the continuous response case. We use the same neural network architecture as before, i.e., the SNN and FNN models but with a softmax activation function in the output layer. FLM is functional linear logistic regression implemented using \textit{fregre.glm} function from \textit{fda.usc} library. Table \ref{tab2} below shows that SNN-based models perform overall better in both linear and non-linear cases, demonstrating the advantages of neural networks. Note that the SNN model can be modified easily in many ways to improve performance.

\begin{table}[h]
	\caption{Prediction errors (mean squared errors) for binary scalar response.}
	\label{tab2}
	\begin{center}
		\begin{tabular}{ll | rrr}
			\hline
			$q$	&	type	&	SNN	&	FNN	&	FLM		\\
			\hline
			5	&	linear	&	0.069	&	0.075	&	\textbf{0.011}	\\
			9	&	linear	&	\textbf{0.083}	&	0.135	&	0.148	\\
			21	&	linear	&	\textbf{0.105}	&	0.361	&	0.207	\\
			5	&	sin	&	0.072	&	0.067	&	\textbf{0.013}	\\
			9	&	sin	&	\textbf{0.082}	&	0.140	&	0.151	\\
			21	&	sin	&	\textbf{0.170}	&	0.405	&	0.230	\\
			\hline
		\end{tabular}
	\end{center}
\end{table}

\subsection{Functional Response and Functional Predictor}

The functional input is the same as before. To facilitate a comparison between a variety of FDA-based models and neural network based models, we generate the functional output using a combination of concurrent, non-concurrent, linear, and non-linear models. 

\begin{itemize}

\item Concurrent, Linear Case (CL): $ Y(t) = \beta(t)X(t)+e(t)$, where function $\beta(\cdot)$ is generated similar to the previous section and $e(\cdot)$ is a centred gaussian process.

\item Concurrent, Non-linear Case (CNL): $ Y(t) =sin( \beta(t)X(t))+e(t)$, where functions $\beta(\cdot),~e(\cdot)$ are generated similar to the previous section.  

\item Non-concurrent, Linear Case (NCL): $ Y(t) = \int \alpha(t,s)X(s)ds+e(t)$, where functions $\alpha(t,s) =  \beta_1(t)\beta_2(s), $ and $ \beta_1(t) = \Sigma_{k=1}^{q}  (k/q)  \phi_k (t_j ),~ \beta_2(t) = \Sigma_{k=1}^{q}  (q-k+1)/(q)  \phi_k (t_j ) $ and $\phi_k (\cdot)$ represents Fourier basis functions

\item Non-concurrent, Non-linear Case (NCNL): $ Y(t) = sin (\int \alpha(t,s)X(s)ds) +e(t)$.
\end{itemize}

We consider the following two neural network models.

\begin{figure}[H]
 \captionsetup[subfigure]{justification=centering}
 \centering
	\subfloat[SNN model]{\label{fig:SNN_model}\includegraphics[width=0.45\textwidth]{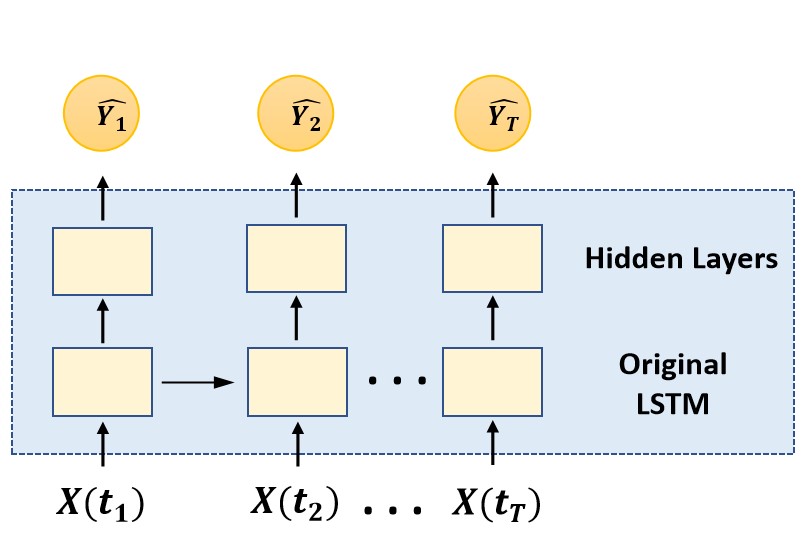}}\quad
	\subfloat[FNN model]{\label{fig:FNN_model}\includegraphics[width=0.45\textwidth]{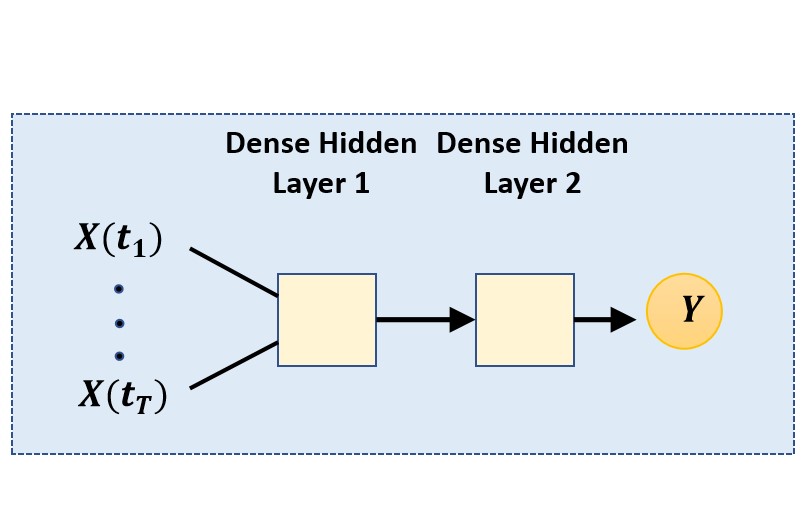}}\\	
        \subfloat[DR model]{\label{fig:DR_model}\includegraphics[width=0.45\textwidth]{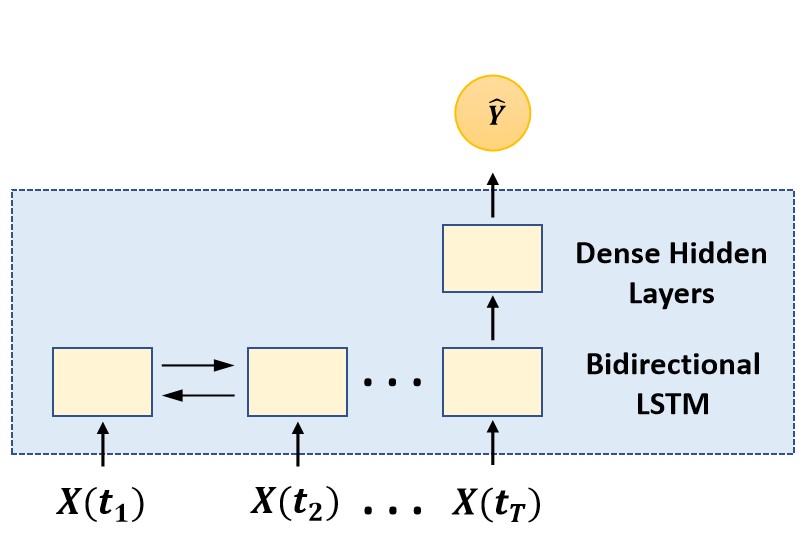}}\quad
	\subfloat[CR model]{\label{fig:CR_model}\includegraphics[width=0.45\textwidth]{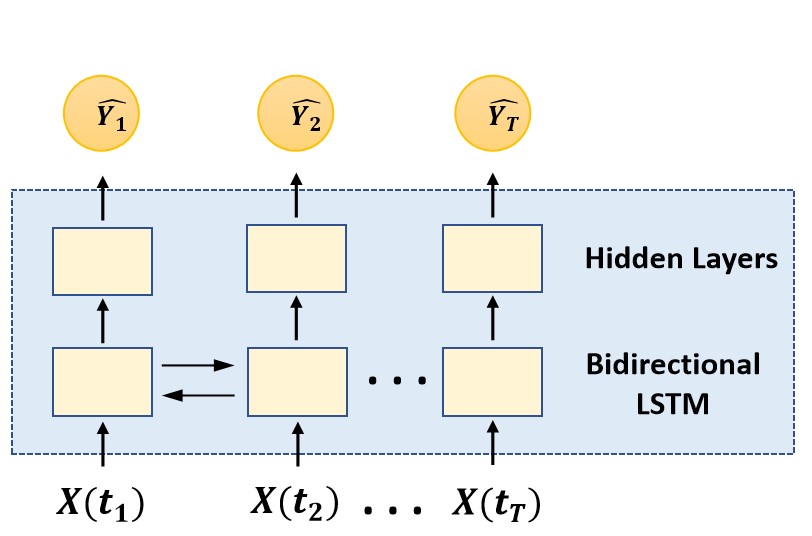}}\\	
	\caption{\textbf{SNN model:} Consisting of an original LSTM layer and a hidden layer, using a many-to-many structure to process sequential data. \textbf{FNN model:}Consisting of two fully connected layers. \textbf{DR model:} Consisting of a bidirectional LSTM layer and a time-distributed layer, using a many-to-one structure to process sequential data. \textbf{CR model:} Consisting of a bidirectional LSTM layer and a hidden layer, using a many-to-many structure to process sequential data.\label{fig4}}
\end{figure}

\textbf{Distributed Response (DR) Model}: This model outputs the predicted response  $\widehat{Y}_j$ at $j^{th}$ unit (or time step) of the sequential network as it processes the input (Figure \ref{fig:DR_model}). We use one bi-directional LSTM layer with 32 units followed by a time distributed dense layer (a dense layer that is repeated at all time points) used with linear activation to obtain the output.

\textbf{Combined Response (CR) Model}: {This model employs a bidirectional LSTM to model the entire input data, and then generates the entire response based on that information. (Figure \ref{fig:CR_model}). The model adopts a many-to-many scheme, taking the output sequence of the bidirectional LSTM as the input of the decoder, and producing the response sequence step by step. The model can fully exploit the bidirectional contextual information of the input data, and ensure the temporal consistency of the response sequence.}

For FDA-based methods, we use a functional linear model based method FLM1 \eqref{m2} from \cite{yao2005functional} that is implemented using \textit{fdapace} library in R. We also implement FLM2, another functional linear model based method using \textit{pffr} function from \textit{refund} library in R \citep{scheipl2016generalized,scheipl2015functional}. The \textit{pffr} function allows implementations of several other functional models. Specifically, we use it to fit a functional non-linear model FNLM1, expressed as $Y(t) = \int_s f(X(s),s,t)+e(t)$. We also use \textit{pffr} function to implement a concurrent model with a smooth varying index effect, referred to as FNLM2 and expressed as $Y(t) =f(X(t),t)+e(t)$, a linear index-varying effect model FLM3, with the formulation  $Y(t) =  \alpha(t) +\beta(t)X(t) +e(t)$, and a non-linear effect model FNLM3 with the expression $Y(t) = f(X(t)) +e(t)$.

{Table \ref{t3} reports the mean squared error of all methods based on 500 replicates. We find that the SNN-based methods DR and CR outperform most of the FDA-based models in terms of prediction performance. In particular, CR significantly surpasses all the FDA-based methods. This advantage is more pronounced in the long sequence experiment with non-concurrent cases. It is worth noting that DR and CR are the most common types in SNN, and their performance can be further enhanced by adjusting the architecture.}


\begin{table}[H]
\centering
\small
\caption{Prediction errors (mean squared errors) for functional response.}
\label{t3}
\resizebox{\linewidth}{!}{
\begin{tabular}{cccccccccc}
\toprule
T & 100 & 100 & 100 & 100 & 20 & 20 & 20 & 20 \\
\cmidrule{2-9}
Case & CL & CNL & NCL & NCNL & CL & CNL & NCL & NCNL \\
\midrule
DR & $6.23\times10^{-3}$ & $5.13\times10^{-3}$ & $2.39\times10^{-3}$ & $2.32\times10^{-3}$ & $2.62\times10^{-2}$ & $1.75\times10^{-2}$ & $1.83\times10^{-2}$ & $2.11\times10^{-2}$ \\
CR & $\mathbf{5.88\times10^{-4}}$ & $\mathbf{4.67\times10^{-4}}$ & $\mathbf{2.16\times10^{-4}}$ & $\mathbf{2.91\times10^{-4}}$ & $\mathbf{1.52\times10^{-4}}$ & $\mathbf{1.48\times10^{-4}}$ & $\mathbf{1.37\times10^{-4}}$ & $\mathbf{1.43\times10^{-4}}$ \\
FNN & $4.24\times10^{-3}$ & $2.77\times10^{-3}$ & $7.28\times10^{-4}$ & $4.33\times10^{-4}$ & $1.06\times10^{-2}$ & $1.06\times10^{-2}$ & $1.06\times10^{-2}$ & $1.06\times10^{-2}$ \\
FLM1 & $8.89\times10^{-1}$ & $6.11\times10^{-1}$ & $2.29\times10^{-1}$ & $1.94\times10^{-1}$ & $2.01\times10^{-1}$ & $2.00\times10^{-1}$ & $2.01\times10^{-1}$ & $1.88\times10^{-1}$ \\
FLM2 & $4.60\times10^{-3}$ & $2.99\times10^{-3}$ & $6.61\times10^{-5}$ & $1.18\times10^{-4}$ & $3.13\times10^{-3}$ & $2.14\times10^{-3}$ & $2.59\times10^{-3}$ & $2.00\times10^{-3}$ \\
FLM3 & $1.07\times10^{-1}$ & $6.78\times10^{-2}$ & $7.13\times10^{-3}$ & $4.11\times10^{-3}$ & $4.21\times10^{-3}$ & $3.67\times10^{-3}$ & $3.87\times10^{-3}$ & $3.52\times10^{-3}$ \\
FNLM1 & $5.35\times10^{-1}$ & $3.05\times10^{-1}$ & $2.48\times10^{-1}$ & $1.66\times10^{-1}$ & $1.11\times10^{-1}$ & $1.09\times10^{-1}$ & $1.08\times10^{-1}$ & $1.06\times10^{-1}$ \\
FNLM2 & $5.35\times10^{-1}$ & $3.05\times10^{-1}$ & $2.36\times10^{-1}$ & $1.64\times10^{-1}$ & $1.16\times10^{-1}$ & $1.12\times10^{-1}$ & $1.12\times10^{-1}$ & $1.08\times10^{-1}$ \\
FNLM3 & $5.35\times10^{-1}$ & $3.05\times10^{-1}$ & $2.24\times10^{-1}$ & $1.57\times10^{-1}$ & $1.12\times10^{-1}$ & $1.39\times10^{-1}$ & $1.39\times10^{-1}$ & $1.35\times10^{-1}$ \\
\bottomrule
\end{tabular}}
\end{table}

\section{Real-World Application}
\label{sec:Appl}
{To validate the conclusions we obtained from the simulation experiments, we conducted data analyses on two real data sets. We compared the prediction performance of FDA and neural network methods in realistic scenarios. The first data analysis utilized the Tecator dataset\citep{reference1}, which consists of a scalar response variable and a functional predictor variable. The second data analysis utilized the Aemet dataset\citep{reference2}, which comprises both predictor variables and response variables related to weather conditions. We compared the prediction errors of different methods.}


\subsection{Tecator Data: Scalar on Function Regression Dataset}
{The Tecator dataset is a commonly used example for functional regression with scalar responses. It encompasses absorbance data across 100 channels from 215 samples of minced meat, along with the proportions of fat, moisture, and protein, determined through professional chemical analysis.
The input data of absorbance captures discrete points within a certain wavelength range and is organized in a matrix.
Our objective is to utilize near-infrared absorbance spectra to predict the fat content in minced meat samples, which is a scalar value.}

To evaluate the predictive performance of different methods on functional regression problems with scalar responses, we conducted the same analysis as the simulation experiment on the Tecator dataset. We divided the dataset into a training set of size 200 and a test set of size 15, and repeated the experiment ten times, taking the mean of the results. Figure \ref{fig:tecator} shows the prediction errors of various methods. Among them, SNN and FNN methods are consistent with the methods introduced in Section 4.1.1. It can be seen that the FLM1 method based on FDA performs the worst, while the SNN and FNN methods based on neural networks are significantly better than all methods based on FDA. Among all methods, SNN method has the lowest prediction error. This indicates that SNN method can capture the features and changes in complex function data more effectively than FDA method, indicating that neural network methods have stronger generalization ability when dealing with complex function data. Compared with FNN method, SNN method can better utilize the temporal features in function data, thereby improving prediction accuracy.

\subsection{Aemet Data: Function on Function Regression Dataset}
{The AEMET dataset refers to the AEMET OpenData dataset obtained from the Spanish Meteorological Agency. This dataset is accessible through the R package. It includes geographical information for each weather station and the average daily values of temperature, precipitation, and wind speed spanning from 1980 to 2009. Specifically, we utilize temperature data to predict logarithmic precipitation data. Both the input and output data are presented in a matrix format with 73 rows and 365 columns.}

We adopted the same analysis method as the simulation experiment in Section 4.2 to conduct a real data experiment on functional regression problems with functional responses, using the Aemet dataset, which contains function variables and response variables related to weather conditions. We divided the dataset into a training set of size 65 and a test set of size 8, and repeated the experiment ten times, taking the mean of the results. 

\begin{figure}[H]
 \captionsetup[subfigure]{justification=centering}
 \centering
	\subfloat[Tecator]{\label{fig:tecator}\includegraphics[width=0.45\textwidth]{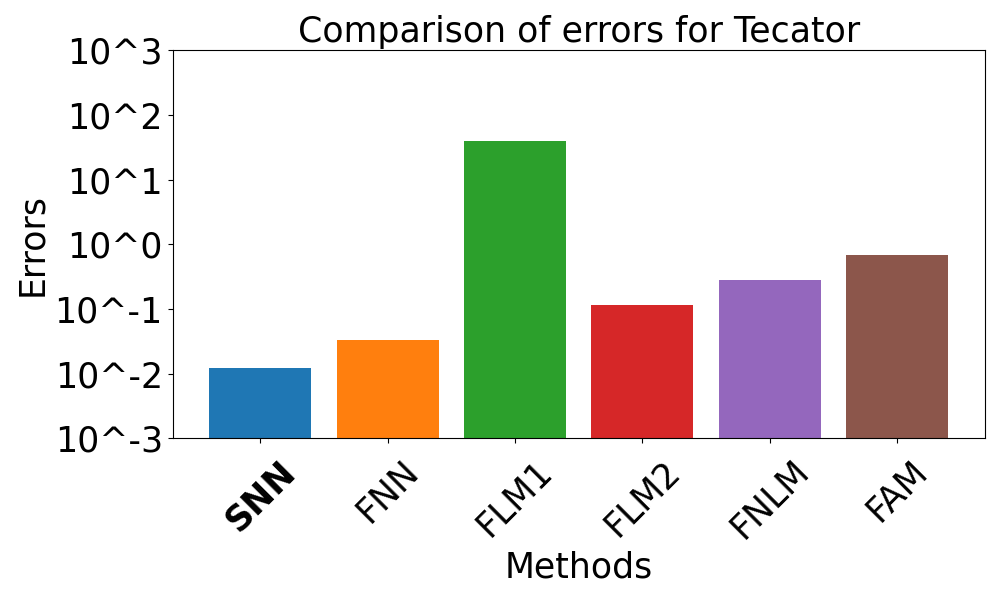}}\quad
	\subfloat[Aemet]{\label{fig:aemet}\includegraphics[width=0.45\textwidth]{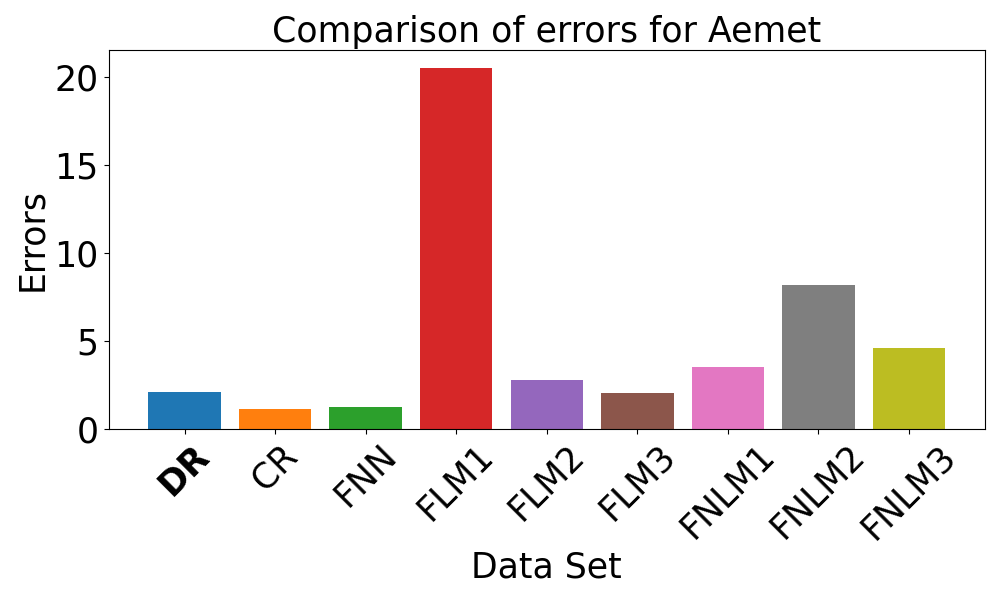}}\\	
	\caption{Comparison of prediction errors for different methods on the Tecator and Aemet Datasets.}\label{fig:data}
\end{figure}

Figure \ref{fig:aemet} shows the prediction errors of various methods. Consistent with the results in Section 5.1, the FLM1 method based on FDA performs the worst. In contrast, the DR, CR and FNN methods based on neural networks are significantly better than all methods based on FDA in terms of prediction performance. This indicates that FDA methods have a significant increase in prediction error when dealing with high-complexity function data, and may not be able to effectively capture the intrinsic structure and dynamics of function data. Among the methods based on neural networks, the CR model using a many-to-many scheme achieves the best prediction effect, because the many-to-many scheme can better handle the alignment relationship between input and output sequences. That is, it can take into account the relevance between elements in the same position of input and output sequences, rather than predicting the entire output sequence as a whole based on the entire input sequence. In this way, it can more accurately capture the changes and patterns in function data, thereby improving prediction performance. In addition, the CR model can adapt to input and output sequences of different lengths or misalignments, while the DR model using a many-to-one scheme requires fixing or padding input sequences, that is, padding all sequences with 0 to be as long as the longest sequence.



\section{Conclusion}
\label{sec:conc}

This paper systematically investigates the application of sequential neural networks in the field of function data analysis and comprehensively compares them with traditional function data analysis methods. Through rigorous numerical experiments and real-world data applications, the study effectively demonstrates the advantages of sequential neural networks in handling high-dimensional and complex data structures in the context of function data, as well as their efficiency in tasks such as prediction and classification. Compared to popular function data analysis regression models, the architecture of sequential neural networks provides greater flexibility, scalability, and effectiveness. It is worth emphasizing that the sequential neural network model proposed in this paper can easily replace the original LSTM with bidirectional LSTM and other variants, and flexibly adjust the activation function types, the number of neural nodes, and the depth of the network to meet the specific requirements of different scenarios.

This study offers novel insights and directions for future research in function data analysis. The sequential neural network demonstrates powerful and flexible characteristics, making it a potent tool for efficient function data processing with broad application potential. We hope this research inspires scholars to explore and innovate in the application of sequential neural networks for function data analysis, further advancing the field and achieving remarkable results in both academia and practical applications.







\bibliographystyle{plainnat}
\bibliography{mybibfile.bbl}
\end{document}